\documentclass{article}

\usepackage{arxiv}

\usepackage[utf8]{inputenc} % allow utf-8 input
\usepackage[T1]{fontenc}    % use 8-bit T1 fonts
\usepackage{hyperref}       % hyperlinks
\usepackage{url}            % simple URL typesetting
\usepackage{booktabs}       % professional-quality tables
\usepackage{amsfonts}       % blackboard math symbols
\usepackage{nicefrac}       % compact symbols for 1/2, etc.
\usepackage{microtype}      % microtypography
\usepackage{lipsum}
\usepackage{amsmath}
\usepackage{graphicx}
\usepackage{float}
\usepackage{eso-pic}
\usepackage{authblk} % 推荐用这个宏包
\usepackage{amsthm}
\usepackage{xcolor}
\usepackage{enumitem}
\newlist{plainlist}{itemize}{1}
\setlist[plainlist]{label={},left=0pt,itemsep=0.65\baselineskip,topsep=0.4\baselineskip}
\usepackage{xcolor}
\usepackage{hyperref} % 若不想要链接颜色，可加 [hidelinks]

\usepackage[numbers]{natbib}
\usepackage{amsthm} % 放在导言区

\newtheorem*{definition}{Definition}

\definecolor{whaleTeal}{RGB}{0,84,109}

\graphicspath{{./images/}}

\title{Why Mask Diffusion does not Work}

\makeatletter
\renewcommand{\@author}{%
	\vspace{-1.2cm}% 整个块往上提 0.8cm
	\begin{tabular}{@{}c@{}}
		\raisebox{-0.2\height}{\includegraphics[height=2cm]{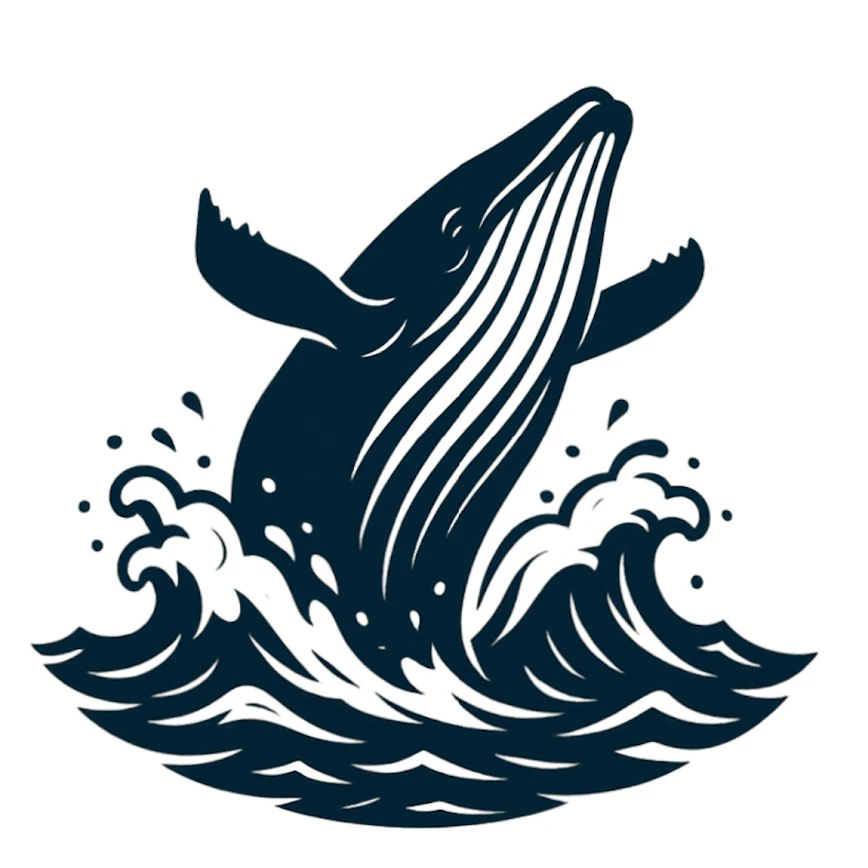}}%
		\hspace{0.6em}{\Large\bfseries WhaleTech.ai}\\[0.4em]
		\texttt{info@whaletech.ai}
	\end{tabular}%
}
\makeatother
\date{} % 不显示日期
\date{}
\renewcommand{\today}{} 
\begin{document}

\maketitle

\begin{abstract}
The main advantages of diffusion language models over autoregressive (AR) models lie in their ability to support parallel generation and bidirectional attention, enabling a more controllable generation process. In recent years, open-source mask diffusion language models have emerged, most of which are based on a variant known as absorbing diffusion. However, this paper demonstrates why mask diffusion faces inherent difficulties in achieving parallel generation and bidirectional attention. We also propose the most effective training and inference strategies for mask diffusion.
\end{abstract}

% keywords can be removed

\section{Introduction}
Diffusion Language Models (DLMs) are rapidly emerging as a powerful and promising alternative to the dominant autoregressive (AR) paradigm in natural language processing. While AR models excel at a wide range of tasks, their sequential, token-by-token generation process imposes a major bottleneck on inference speed and computational efficiency. In contrast, DLMs generate tokens in parallel through an iterative denoising process. This core mechanism provides inherent advantages in reducing inference latency and naturally incorporating bidirectional context, making DLMs a compelling choice for addressing the trade-off between generation quality and speed.

To adapt diffusion for discrete language data, one of the primary directions has been the development of continuous-space models \citep{li2022diffusionlm,dieleman2022continuous,han2023ssdlm}. This approach first maps discrete tokens into a continuous embedding space where the denoising process is performed. The forward process gradually transforms the token embeddings into noise, and the reverse process learns to invert this corruption, starting from noise and progressively generating a clean embedding. After a denoised embedding is generated, a final rounding step maps it back to a discrete token from the vocabulary. 

In recent years, discrete-space DLMs have gained increasing popularity as they define the diffusion process directly on the vocabulary of tokens, avoiding the need for a continuous embedding space during diffusion itself. In this paradigm, the forward process corrupts a sequence by applying a transition matrix at each step, which defines the probability of a token transitioning to any other token \citep{austin2023structured,lou2024discrete,hoogeboom2021argmax}. The model then learns to reverse these transitions by predicting the original token distribution given the corrupted sequence. 

Masked Diffusion Models represent a modern and highly effective evolution of discrete diffusion language models, forming the foundation for several recent large-scale efforts such as DiffuLLaMA, LLaDA and Dream \citep{ gong2025scaling,nie2025large, ye2025dream}. The core principle involves a forward process where tokens in a sequence are probabilistically replaced with a special \texttt{[MASK]} token \citep{austin2023structured}. During training, the model learns to predict the original content for these \texttt{[MASK]} positions, with the loss computed only over the \texttt{[MASK]} tokens. The generation process starts with a fully \texttt{[MASK]} sequence. In each iterative step, the model predicts several tokens. This iterative refinement continues until all \texttt{[MASK]} tokens are resolved.

In our work, we demonstrate that mask diffusion language models face inherent difficulties in achieving true parallel generation and bidirectional attention, due to the following reasons:
\begin{enumerate}
	\item The model outputs the conditional marginal distribution for each \texttt{[MASK]} token, rather than the joint probability over all \texttt{[MASK]} tokens. This implies that parallel sampling cannot be theoretically guaranteed.
	\item The distributions over \texttt{[MASK]} tokens distant from the unmasked positions are often smooth, and their most probable token IDs tend to be homogeneous. This essentially implies that, although many of the probabilities produced by the model are correct, they provide little useful information for sampling.
	\item For mask diffusion language model, the most reliable and stable generation strategy is likely still the autoregressive (AR) approach, which makes it difficult to effectively leverage bidirectional attention during the generation process.
\end{enumerate}

\section{Preliminaries}
\subsection{Auto-regressive Modeling}
Given a token sequence $x^{1:N} = (x^1, x^2, \dots, x^N)$, an autoregressive (AR) language model 
factorizes the joint probability into a product of conditional probabilities:
\begin{equation}
p(x^{1:N}) = \prod_{n=1}^{N} p(x^n \mid x^{<n}),
\end{equation}
where $x^{<n} = (x^1, \dots, x^{n-1})$ denotes the prefix before position $n$. 
This left-to-right decomposition ensures that each token is generated conditioned on all previously observed tokens.

During training, the standard paradigm is to maximize the log-likelihood of the observed sequence, 
which is equivalent to minimizing the negative log-likelihood (NLL):
\begin{equation}
\mathcal{L}_{AR} = - \sum_{n=1}^{N} \log p_\theta(x^n \mid x^{<n}),
\end{equation}
where $\theta$ denotes the model parameters. 
In practice, this loss reduces to the token-level cross-entropy between the predicted distribution and the ground-truth token at each position.

\subsection{Mask Diffusion Modeling}

Let $x_0 \in \Delta_K$ be a one-hot token (with the $K$-th category \texttt{[MASK]} denoted by $m$).
Mask diffusion defines a forward process that interpolates between the clean $x_0$ and the \texttt{[MASK]} token
via a monotone schedule $\alpha_t \in [0,1]$ ($\alpha_0=1,\alpha_1=0,t\in [0,1]$):
\begin{equation}
	q(x_t \mid x_0) = \mathrm{Cat}\left(x_t;\; \alpha_t x_0 + (1-\alpha_t) m\right).
\end{equation}

For $s<t$, the transition is
\begin{equation}
	q(x_t \mid x_s) = \mathrm{Cat}\left(x_t;\; \alpha_{t|s} x_s + (1-\alpha_{t|s}) m\right),
	\quad \alpha_{t|s}=\frac{\alpha_t}{\alpha_s}.
\end{equation}

For absorbing diffusion ($\pi=m$), the posterior is
\begin{equation}
	q(x_s \mid x_t, x_0) =
	\left\{
	\begin{array}{ll}
	\mathrm{Cat}(x_s; x_t), & \ \ \ x_t \neq m, \\[6pt]
	\mathrm{Cat}\left(x_s;\; \frac{(1-\alpha_s)m + (\alpha_s-\alpha_t)x_0}{1-\alpha_t}\right), & \ \ \ x_t = m.
	\end{array}
	\right.
\end{equation}

We approximate the reverse posterior by a denoiser $f_\theta(x_t,t)\in\Delta_K$:
\begin{equation}
	p_\theta(x_s \mid x_t) =
	\left\{
	\begin{array}{ll}
	\mathrm{Cat}(x_s; x_t), & \ \ \ x_t \neq m, \\[6pt]
	\mathrm{Cat}\left(x_s;\; \frac{(1-\alpha_s)m + (\alpha_s-\alpha_t)f_\theta(x_t)}{1-\alpha_t}\right), & \ \ \ x_t = m,
	\end{array}
	\right.
\label{eq:weight}
\end{equation}

where $\langle f_\theta(x_t),m\rangle=0$ \citep{shi2025simplified}.

\begin{equation}
\mathrm{KL}\!\left(q(x_s \mid x_t, x_0)\,\|\,p_\theta(x_s \mid x_t)\right)
=
\left\{
\begin{array}{ll}
\frac{\alpha_s - \alpha_t}{1 - \alpha_t} \,
\mathrm{KL}\!\left(x_0 \,\|\, f_\theta(x_t)\right), & x_t = m, \\[6pt]
0, & x_t \neq m .
\end{array}
\right.
\end{equation}

\paragraph{ELBO.} The variational evidence lower bound is
\begin{equation}
	 \mathcal{E} = \mathbb{E}_{q}\!\left[ \log p_\theta(x_0 \mid x_{t(1)}) \right] - \sum_{i=2}^{T} \mathbb{E}_{q}\!\left[ \mathrm{KL}\!\left( q\!\left(x_{s(i)} \mid x_{t(i)}, x_0\right) \,\|\, p_\theta\!\left(x_{s(i)} \mid x_{t(i)}\right) \right) \right] - \mathbb{E}_{q}\!\left[ \mathrm{KL}\!\left( q\!\left(x_{t(T)} \mid x_0\right) \,\|\, p\!\left(x_{t(T)}\right) \right) \right]. 
\end{equation}
% --- NELBO = training objective --- 
\paragraph{NELBO (training objective).} Equivalently, the training loss as the negative ELBO is 
\begin{equation}
	\mathcal{L}_{\mathrm{NELBO}} = - \mathbb{E}_{q}\!\left[ \log p_\theta(x_0 \mid x_{t(1)}) \right] + \sum_{i=2}^{T} \mathbb{E}_{q}\!\left[ \mathrm{KL}\!\left( q\!\left(x_{s(i)} \mid x_{t(i)}, x_0\right) \,\|\, p_\theta\!\left(x_{s(i)} \mid x_{t(i)}\right) \right) \right] + \mathbb{E}_{q}\!\left[ \mathrm{KL}\!\left( q\!\left(x_{t(T)} \mid x_0\right) \,\|\, p\!\left(x_{t(T)}\right) \right) \right].
\end{equation}  

Finally,

\begin{equation}
\mathcal{L}
= \sum_{i=1}^{T} 
\left( 
- \frac{\alpha_{s(i)} - \alpha_{t(i)}}{1-\alpha_{t(i)}} \,
1_{\{x_{t(i)}=m\}}\, x_0^\top \log f_\theta(x_{t(i)})\,
\right),
\end{equation}

where $s(i+1)=t(i)=\frac{i}{T}$, $x_0$ is a one-hot vector. 

In the continuous time limit where $T \to \infty$ \citep{shi2025simplified}, we set a small timestep 
$\Delta_t = t(i) - s(i) = \frac{1}{T} \in (0,1)$. 
The sum over timesteps becomes an integral, and we have a limit
\begin{equation}
	\alpha_t' = \frac{\alpha_{t(i)} - \alpha_{s(i)}}{t(i) - s(i)},
\end{equation}

and 
\begin{equation}
\mathcal{L}
= \int_{0}^{1}
 \frac{\alpha_{t'}}{1-\alpha_{t}} \,
1_{\{x_t=m\}}\, x_0^\top \log f_\theta(x_t)\,
dt
.
\end{equation}

In practice, given a token sequence $x_0^{1:N} = (x_0^1, x_0^2, \dots, x_0^N)$ and a time moment $t\in (0,1]$:

\begin{equation}
	\mathcal{L}
	=\frac{\alpha_{t'}}{1-\alpha_{t}} \,
	\sum_{j=1}^{N}
	\left(
	1_{\{x_t^j=m\}}\, (x_0^j)^\top \log f_\theta(x_t^{1:N})
	\right)
	.
\label{eq:loss}
\end{equation}

\section{Challenges of Mask Diffusion: A Theoretical Perspective}
To better understand the limitations of mask diffusion, this section provides a theoretical analysis of why it struggles to support parallel generation and bidirectional attention.

\subsection{Theoretical Setup}
\label{assum}
For our theoretical analysis, we adopt the following idealized assumptions:

\noindent \textbf{Assumption 1 (\texttt{[MASK]} token neutrality).} 
The \texttt{[MASK]} token itself carries no intrinsic information (except location information which we don't predict) and can thus be omitted when formulating conditional probabilities. That means the prediction of a \texttt{[MASK]} token only relies on unmasked tokens.

\medskip
\noindent \textbf{Assumption 2 (Length independence).} 
Although performance in practice depends on the length of given \texttt{[MASK]} sequence (for example, variations in the reserved length may substantially affect the length of the model’s valid output), we abstract away from this factor and assume that diffusion model has already learned the complete data distribution under any given truncation length (equivalently, one may regard the model as reserving an infinitely long sequence of \texttt{[MASK]} tokens.). Consequently, the model’s output is unaffected by the total sequence length: its prediction on a \texttt{[MASK]} token is determined solely by the set of currently unmasked tokens. 

\subsection{Marginal Distribution Analysis}
In this section, we reveal why the model's output should be regarded as a marginal distribution and the distance-dependent features of this distribution.

\subsubsection{Weighted Loss Function}
We will explain why, from a theoretical perspective, the weights in the weighted loss function are not essential, although they can provide practical benefits in training.

The loss function (Equation~\ref{eq:loss}) essentially measures the model’s ability to reconstruct the original data from the noised input at any time step $t$. The only explicit dependence on $t$ appears in the weighting term (e.g., $\frac{\alpha_{t'}}{1-\alpha_{t}}$). Theoretically, however, if the model is sufficiently trained (all loss equal to 0), these coefficients are not crucial. The role of the weights is mainly to encourage the model to focus more on cases with small $t$, while down-weighting the loss at large $t$, thereby stabilizing the training curve\citep{shi2025simplified}.

In fact, the weight comes from Equation~\ref{eq:weight}; Since $f_\theta(x_t)$ has zero components at the positions corresponding to \texttt{[MASK]} tokens, we need weighted combination of $f_\theta(x_t)$ and $m$ to approximate $q(x_s|x_t)$. That actually means the model $f_\theta$ only focuses on predicting the probabilities of unmasked tokens, one step back to the initial state $(t=0)$.

Therefore, the model does not care about the exact timestep it is in during inference (unlike diffusion for image generation, timestep matters); at different timesteps, two sequences may well be in the same noise state, leading the model to produce identical outputs.

\subsubsection{Model Output as Marginal Distributions}
In discrete settings, it is necessary to manually define neighborhoods, as in concrete score matching\citep{meng2023concrete}. Mask diffusion is essentially a single-token transition process, similar to SEDD\citep{lou2024discrete}. Since noise is added independently to each token, The forward process can be regarded as many independent single-token transitions executed in parallel. This, in turn, implies that the reverse denoising process is also fundamentally a single-token transition problem: the model learns how to update one token at a time. However, when multiple tokens are updated simultaneously, there is no guarantee of their mutual coherence. We will demonstrate it in the following example.

\paragraph{A toy experiment}
Consider a toy dataset consisting of 100 sequences, each of length five tokens. All sequences share the same middle subsequence CD, and the final token E is unique across sequences (i.e., each sequence has a distinct value of E). The first two tokens pair has four cases: 34 sequences begin with $AB$, 21 with $AB'$, 35 with $A'B$, and 10 with $A'B'$. Apply mask diffusion to this toy dataset and assume the model is sufficiently trained. If the input sequence is $\texttt{[MASK]}\texttt{[MASK]}CD\texttt{[MASK]}$, what would the model’s outputs on the \texttt{[MASK]} tokens be?

In the above experiment, the model fails to retain the joint probability of the first token pair. Instead, it only output $P(A|CD)=0.55$ and $P(A'|CD)=0.45$ at the first \texttt{[MASK]} position and $P(B|CD)=0.69$ and $P(B'|CD)=0.31$ at the second \texttt{[MASK]} position (Under Assumption 1, we omit the explicit form of the \texttt{[MASK]} tokens when computing conditional probabilities). If now you decide to predict the first two tokens at once, your choice will be $AB$ while $A'B$ is the most probable pair.  

This experiment highlights a problem. Although the model is exposed to many token combinations as supervision signals during training, it only outputs \textbf{the marginal distribution of each individual \texttt{[MASK]} token}, unlike image generation diffusion, which directly learns the gradient of the log joint probability.

\begin{figure}[H] % picture
	\centering
	\includegraphics[width=0.8\textwidth]{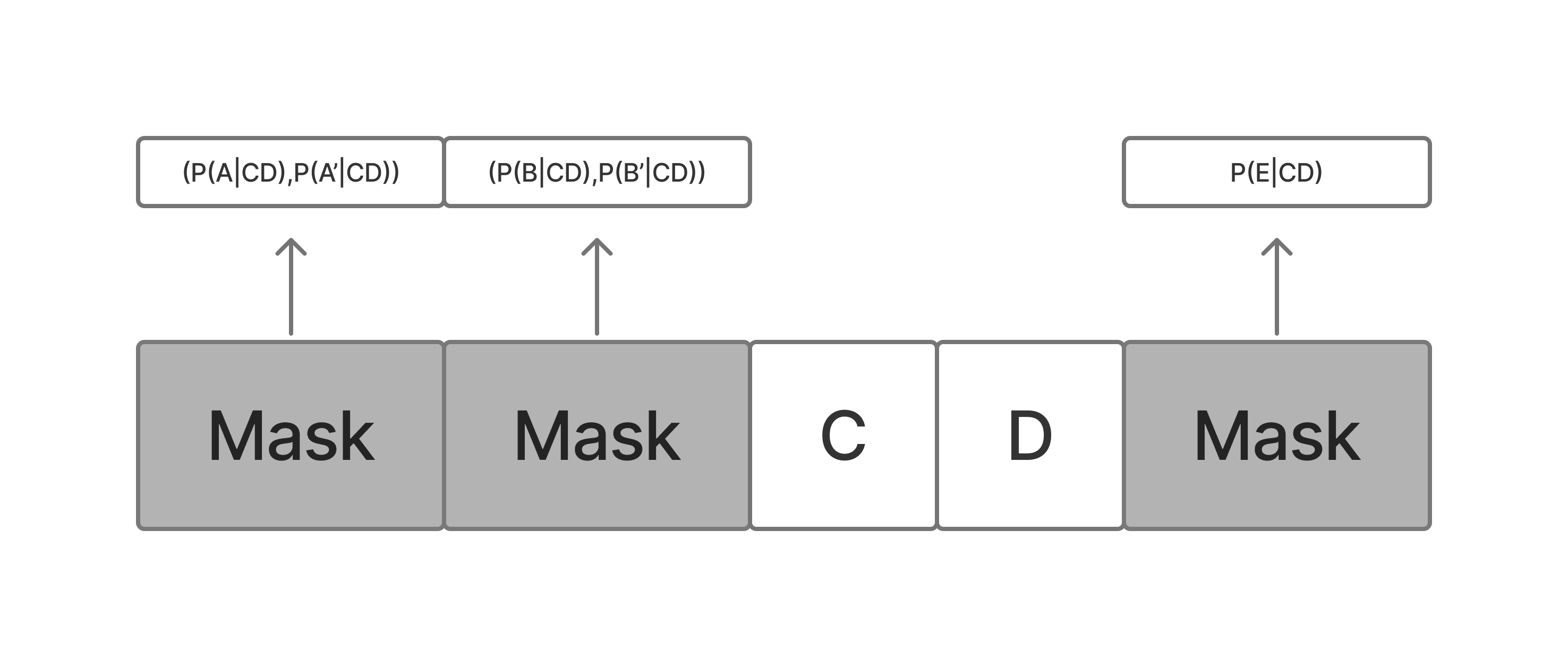}
	\caption{Model Output as Marginal Distributions}
	\label{fig:fig0}
\end{figure}

As a result, similar to autoregressive models, mask diffusion model cannot guarantee maximization of the joint probability. You may need auxiliary techniques, such as beam search, to find token combinations that maximize the joint probability.

\subsubsection{Marginal Distributions as a Function of Distance}
In general, the marginal distribution of a \texttt{[MASK]} token tends to be sharper when it is closer to unmasked tokens, whereas tokens farther away exhibit smoother distributions. This phenomenon arises because the conditional dependency on observed tokens weakens with distance, leading to greater uncertainty in the predicted probabilities.

This phenomenon can be described through a set of mathematical assumptions.

\noindent \textbf{Assumption 3 .} 
Under autoregressive (AR) conditions, the prediction of the next token is characterized by a Zipfian distribution. In formal mathematical expression, given all preceding unmasked tokens $T_1,T_2,...,T_n$, the next token prediction $P(T_{n+1}|T_nT_{n-1}...T_1)$ is a Zipfian distribution \footnote{$P(x=n)=\frac{n^{-s}}{\sum_{k=1}^{N} k^{-s}}, \quad n = 1,2,\dots,N$}\citep{Piantadosi2014Zipf}. 

\noindent \textbf{Assumption 4 .} 
Given all preceding unmasked tokens $T_{n-1},T_{n-2},...T_{1}$. Different choices of $T_{n}$ will lead to noticeable differences in the prediction of next token $T_{n+1}$. In formal mathematical expression, the peak token IDs of $P(T_{n+1}|T_nT_{n-1}...T_1)$ will be different for any two distinct $T_n$. Equivalently, the peak token positions constitute a permutation matrix if $T_{n+1}$ and $T_n$ vary within the range of unmasked tokens.

\paragraph{Remark.}
Both assumptions are in fact rather strong, and they may not necessarily hold locally. However, in the sense of expectation, this trend exists.

Under these assumptions, we can demonstrate the existence of the aforementioned phenomenon.

Given a prompt (which may, for instance, be a user’s query), the model reserves an infinitely long sequence of \texttt{[MASK]} tokens (Assumption 2); From left to right, we denote them as $T_1,T_2,...,T_n,...$, random variables initialized at the \texttt{[MASK]} tokens. As previously suggested, the model outputs marginal distributions of all \texttt{[MASK]} tokens. 

\begin{figure}[H] % picture
	\centering
	\includegraphics[width=0.8\textwidth]{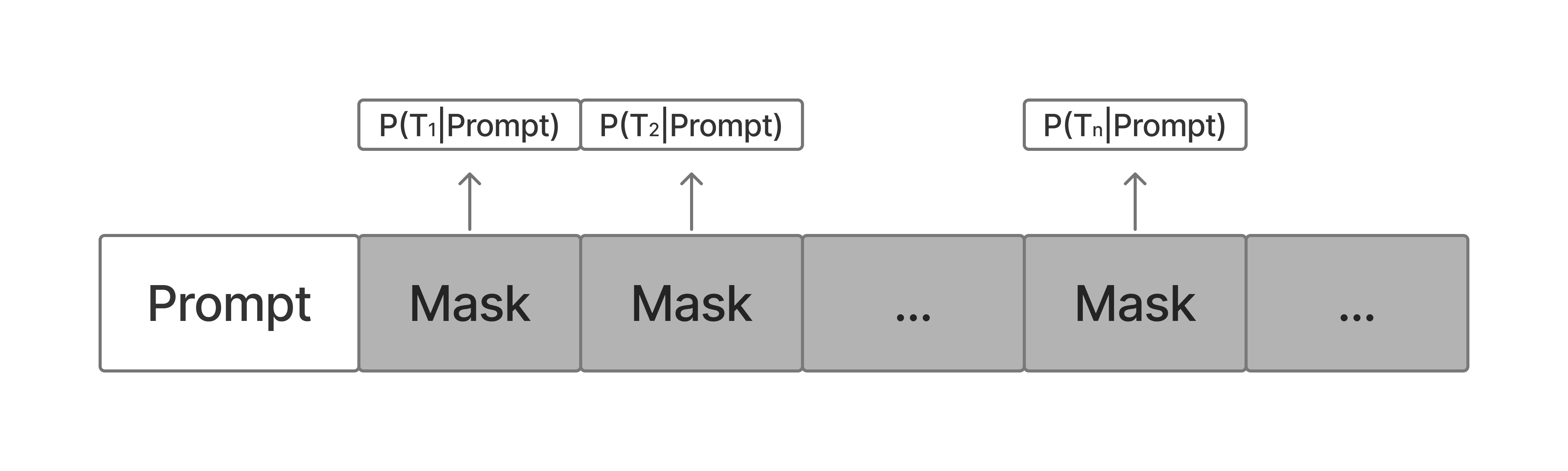}
	\caption{Initial state during inference}
	\label{fig:fig3}
\end{figure}

Then $P(T_n|T_{n-1}...T_1,prompt)$ is a Zipfian distribution for all $n\ge 1$. We assume that they share the same parameters $N$ and $s$, thus leading to the same top1 prob $\omega_1$ and top2 prob $\omega_2$ \footnote{$\omega_1=\frac{1}{\sum_{k=1}^{N} k^{-s}}, \omega_2=\frac{2^{-s}}{\sum_{k=1}^{N} k^{-s}}$}.

We know 
\begin{equation}
\begin{aligned}
&P(T_n|prompt) \\
=&\sum_{T_1}\cdots \sum_{T_{n-2}}\sum_{T_{n-1}}
P(T_n|T_{n-1}\cdots T_1,prompt)P(T_{n-1}|T_{n-2}\cdots T_1,prompt)\cdots P(T_1|prompt),
\end{aligned}
\end{equation}

where
\begin{equation}
\label{eq:15}
\begin{aligned}
&\sum_{T_{n-1}}P(T_n|T_{n-1}\cdots T_1,prompt)P(T_{n-1}|T_{n-2}\cdots T_1,prompt)\\
\le & \omega_1 \text{Max}P(T_{n-1}|T_{n-2}\cdots T_1,prompt)+
\omega_2 (1-\text{Max}P(T_{n-1}|T_{n-2}\cdots T_1,prompt)) \quad \text{(Assumption 4)} \\
= & (\omega_1-\omega_2)\text{Max}P(T_{n-1}|T_{n-2}\cdots T_1,prompt)) +\omega_2.
\end{aligned}
\end{equation}

In addition,
\begin{equation}
\begin{aligned}
& \sum_{T_{n-2}}\Big[
(\omega_1-\omega_2)\text{Max}P(T_{n-1}|T_{n-2}\cdots T_1,prompt)+\omega_2
\Big]P(T_{n-2}|T_{n-3}\cdots T_1,prompt) \\
\le & (\omega_1-\omega_2)\Big[
(\omega_1-\omega_2)\text{Max}P(T_{n-2}|T_{n-3}\cdots T_1,prompt)+\omega_2
\Big]+\omega_2. \quad \text{(As Equation~\ref{eq:15})}
\end{aligned}
\end{equation}

Through iteration,
\begin{equation}
\label{eq:17}
\begin{aligned}
&P(T_n|prompt) \\
\le & (\omega_1-\omega_2)^{n-1}\text{Max} P(T_1|prompt)+\omega_2 \frac{1-(\omega_1-\omega_2)^{n-1}}{1-(\omega_1-\omega_2)} \\
= & (\omega_1-\omega_2)^{n-1}\omega_1+\omega_2 \frac{1-(\omega_1-\omega_2)^{n-1}}{1-(\omega_1-\omega_2)} \quad \text{(Assumption 3)} \\
= & \frac{\omega_2}{1-(\omega_1-\omega_2)}+(\omega_1-\omega_2)^{n}\frac{1-\omega_1}{1-(\omega_1-\omega_2)}.
\end{aligned}
\end{equation}

As $n$ increases, this upper bound converges to
\begin{equation}
	\frac{\omega_2}{1-(\omega_1-\omega_2)}.
\end{equation}

Although this is a coarse upper-bound estimate, it can still provide us with some intuition: the maximum probability will decay toward a lower bound which is small but not zero.

\subsubsection{Experiment}
We use LLaDA as the experiment model and set the maximum sequence length to 128. We consider four scenarios; In the first scenario, we design 100 simple questions; in the second, 100 moderately complex mathematics problems; in the third, 100 nonsensical questions; and in the fourth, 100 questions from the domain of humanities and the arts. We recorded the model's average maximum probability at all position of \texttt{[MASK]} tokens. Meanwhile, we set $s=2.31$, $N=130000$, under which you will observe the upper-bound curve obtained from our own estimation. 

\begin{figure}[H] % picture
	\centering
	\includegraphics[width=0.9\textwidth]{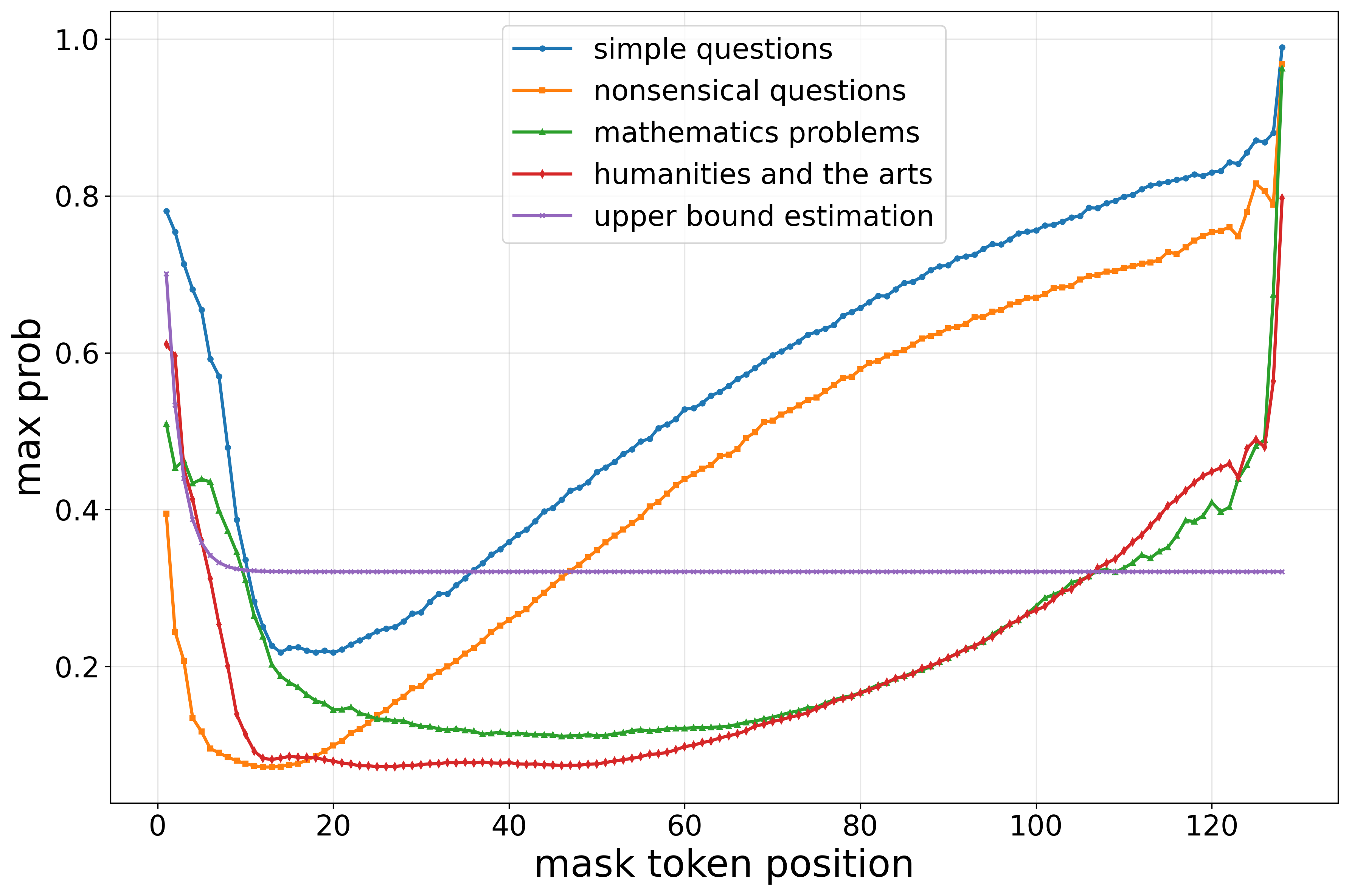}
	\caption{True average max prob and upper bound estimation}
	\label{fig:fig4}
\end{figure}

Regarding the experimental results, we have made several interesting observations.

\begin{description}
	\item[Observation 1 (Increasing Maximum Probability).] 
	In practice, one may observe that the maximum probability component of \texttt{[MASK]} tokens near the end of a sentence tends to be relatively large. This phenomenon can often be attributed to the presence of special tokens such as end-of-text (or similar tokens indicating sequence termination). If such tokens occur frequently during training, they are more likely to dominate the probability mass during inference. However, this is not necessarily desirable, as the end-of-text token itself carries no substantive meaning, and its premature prediction effectively reduces the space available for meaningful tokens.
	\item[Observation 2 (Similarities and Differences).]  All the curves exhibit a rapid decline within the first ~15 tokens, with the differences lying in their lower bounds and the positions at which they begin to rise. When the model exhibits a preference for shorter responses, the upward turning point of the curve emerges earlier. Moreover, when the task is both clear and simple, the lower bound of the probability distribution maybe higher.
	\item[Observation 3 (Ground Truth vs. Prediction).] Our estimation follows the same trend as the actual behavior in the beginning, however, the true average maximum probability usually converges to a smaller value (but still noticeably greater than zero). Our predicted curve does not rise afterwards, since our assumption does not account for length constraints or the special end-of-text token.
\end{description}

Overall, a smaller maximum probability in fact implies that the probability distribution becomes smoother and less favorable for sampling. 

\paragraph{Homogenization of Distant Mask Predictions}Some tokens are function words, which occur very frequently and can combine with many different tokens. In contrast, meaningful tokens occur less frequently and have more limited patterns of combination. This implies that function tokens are more likely to absorb probability mass transferred from the preceding token, which makes them more likely to dominate the probability distribution at positions farther away from unmasked tokens. This explains why, when all \texttt{[MASK]} tokens are predicted simultaneously, positions far from unmasked tokens often collapse to repeated function words such as "," or other high-frequency tokens.

\begin{table}[h]
	\centering
	\caption{Homogenization of Distant Mask Predictions}
	\resizebox{\textwidth}{!}{%
		\begin{tabular}{|c|c|c|c|c|c|c|c|c|c|}
			\hline
			I & plays & the & have & role & in & perceive & reality & . & perceive \\ \hline
			the & . & . & the & and & the & and & and & and & and \\ \hline
			and & , & and & and & , & , & and & and & , & , \\ \hline
			, & , & , & , & , & , & , & , & , & , \\ \hline
			, & , & , & , & , & , & , & , & , & , \\ \hline
			, & , & , & , & , & , & , & , & , & , \\ \hline
			, & , & , & , & , & , & , & , & , & , \\ \hline
			, & , & , & , & , & , & , & , & , & , \\ \hline
			, & , & , & , & , & , & , & , & , & , \\ \hline
			, & , & , & , & , & , & , & , & endoftext & endoftext \\ \hline
			endoftext & endoftext & endoftext & endoftext & endoftext & endoftext & endoftext & endoftext & endoftext & endoftext \\ \hline
			endoftext & endoftext & endoftext & endoftext & endoftext & endoftext & endoftext & endoftext & endoftext & endoftext \\ \hline
			endoftext & endoftext & endoftext & endoftext & endoftext & endoftext &  . & eot\_id  &  &  \\ \hline
		\end{tabular}%
	}
\end{table}

\noindent \textbf{Conjecture.} 
At sufficiently large distances with an infinite length of \texttt{[MASK]} tokens, the distributions become almost identical. With a fixed given length, this near-identical behavior appears in the middle parts of the sequence.

The above characteristics are inevitable phenomena of mask diffusion when the model is sufficiently trained. 

\subsection{Parallel Sampling}
In principle, parallel sampling is infeasible because the assumption of conditional independence does not hold, as also assumed in SEDD (though SEDD allows subsequent adjustments). Due to the distance-dependent behavior of the distribution, marginal conditional probabilities of the initial \texttt{[MASK]} tokens may still exhibit sharp features. Nevertheless, parallel sampling is likely to suffer from reduced joint probability, where unusual token combinations may appear together despite each token being individually probable. 

Given a prompt, we introduce three metrics to evaluate the joint probability of parallel sampling. We assume that we only consider sampling the first few adjacent \texttt{[MASK]} tokens in parallel, as this is the most reliable choice among all parallel sampling strategies.

\begin{description}
	\item[Metric 1 (The upper bound of joint probability).] 
	If you wish to sample the first $n$ \texttt{[MASK]} tokens, and their corresponding maximum probabilities are $p_1,\cdots,p_n$, then the maximum joint probability of the token combination sampled according to these maximum probabilities is $\text{min}(p_1,\cdots,p_n)$.
	\item[Metric 2 (Independence assumption).] If you assume that the selection of each token is mutually independent, then the joint probability is given by $\prod_{i=1}^{n}p_i$.
	\item[Metric 3 (The lower bound).] If you decide to sample the first $n$ \texttt{[MASK]} tokens according to their maximum probabilities, then in the worst case the joint probability is $\text{max}(0,\sum_{i=1}^{n}p_i-(n-1))$\footnote{Bonferroni bound}. 
\end{description}

Although these metrics are rather coarse, they can be used to estimate the upper bound of parallel sampling. 

\begin{definition}
	The $PPL$ of a sequence $(x^1,x^2,\cdots,x^n)$ is defined as $P(x^1,x^2,\cdots,x^n)^{-\frac{1}{n}}$.
\end{definition}

\paragraph{Remark.} 
Approximately speaking, perplexity (PPL) reflects the reciprocal of the average joint probability per token. A larger PPL indicates greater uncertainty for each token, which in turn implies a lower joint probability for the entire sequence.

For example, given a prompt and an infinite sequence of \texttt{[MASK]}tokens; With Equation~\ref{eq:17} and Metric 2, assuming $s=1.05$ and $N=150000$, we obtain a table showing how approximate values of PPL vary with the number of parallel samples. If the upper bound of PPL is set to 17, then the upper bound on parallel sampling is 6.
\begin{table}[h]
	\caption{Approximate Values of PPL}
	\centering
	\begin{tabular}{cc}
		\toprule
		Parallel Samples & PPL \\
		\midrule
		1  &  9.6618 \\  % 在这里填入你的数据
		2  &  13.1995 \\
		3  &  14.8584 \\
		4  &  15.8570 \\
		5  &  16.4716 \\
		6  &  16.9408 \\
		7  &  17.3027 \\
		8  &  17.6066 \\
		\bottomrule
	\end{tabular}
	\label{tab:ppl_parallel}
\end{table}

The real situation is considerably more complex. Considering Homogenization of Distant Mask Predictions, Parallel sampling may lead to the same token being predicted at two adjacent positions, which in turn causes the joint probability to drop sharply. 

Although estimating the upper bound of parallel prediction is challenging-since it depends on different semantic tasks, such a bound does exist and may be smaller than expected.

\paragraph{Remasking techniques}
Although mask diffusion also learns single-token transitions, it differs from SEDD in that the denoising process fixes the unmasked tokens. While certain remasking techniques exist\citep{wang2025remasking}, they can not achieve parallelism, as remasking multiple unmasked tokens simultaneously may cause substantial information loss, nor acceleration, since they require an additional prediction step.

\paragraph{A Random Initialization Strategy Inspired by SEDD}
One approach is to randomly initialize some \texttt{[MASK]} tokens and reveal others using these sparse signals, then remask the initialized tokens. Repeat many times to fill all \texttt{[MASK]}tokens. This is similar to a SEDD-inspired approximation, but it is inefficient(only the outputs of \texttt{[MASK]} tokens matter) and does not significantly improve speed. In addition, the ratio of randomly initialized tokens is important, as it sets the fraction of information the model fully relies on at each step.

\subsection{Generation Order and Training Strategies}
\subsubsection{Generation Order}
The order of generation is crucial, as it can lead to entirely different outputs. We argue that in mask diffusion-if only one token can be predicted at a time and previously unmasked tokens cannot be adjusted-the autoregressive generation order is optimal in general case. 

This view is based on empirical intuition: when only one token can be predicted at a time, the best choice is to select the token with the highest marginal probability. From our previous analysis, we see that, in general, the predicted marginal distributions tend to decrease from left to right. In this case, the most reliable strategy is to choose the token with the largest marginal probability, which is typically the leftmost token. However, once the leftmost \texttt{[MASK]} token has been chosen, the next token often becomes the one with the largest marginal probability at that step. Thus, the generation process is often nearly autoregressive, with occasional instances of locally non-autoregressive behavior.

In this case, the model does not effectively exploit the advantages of bidirectional attention. 

\paragraph{Reverse Order Generation}
If you sample according to confidence, which is also what we would typically do in practice. Occasionally, the generation process appears to proceed from the end backward. This arises from the influence of the end-of-text token, which is itself the most frequent token. When the sequence length is large, the model may assign the highest confidence to generating an end-of-text token at the final positions. Once one such token is generated, it can in turn increase the likelihood of generating another end-of-text token in the preceding position. Overall, this is not desirable, as it prematurely fixes the length of the sentence.

\paragraph{Random Generation}
Random-order generation often produces fluent text. However, it can only generate one token at a time; otherwise, it becomes difficult to ensure coherence among tokens. Moreover, it is difficult to guarantee that the joint probability of such sampling results is maximized.

\subsubsection{Training Strategies}

In mask diffusion, during training we randomly mask a relatively long sequence and attempt to reconstruct it. This requires our training to cover the $2^n-1$ possible scenarios that a sentence may face, which is quite difficult to achieve. However, once we determine that inference is carried out in an almost autoregressive (AR) manner, most of the situations covered during training will not actually occur at inference time, thereby introducing substantial redundancy in training.

We introduce the optimal inference and training method, which is well aligned with the current generation approach.

\paragraph{Semi-AR Generation in Small Blocks}
During inference, we define a small block (e.g., of size 4 or 8) and generate within it using a diffusion process (or parallel sampling with the block size as the upper bound). Once the predictions for the block are completed, we proceed to the next small block, until the model outputs the end-of-text token. This essentially corresponds to semi-autoregression\citep{han2023ssdlm}, but with smaller blocks; the reason is that \texttt{[MASK]} tokens far from the given condition are difficult to exploit, making large block sizes nearly ineffective.

\begin{figure}[H] % picture
	\centering
	\includegraphics[width=0.8\textwidth]{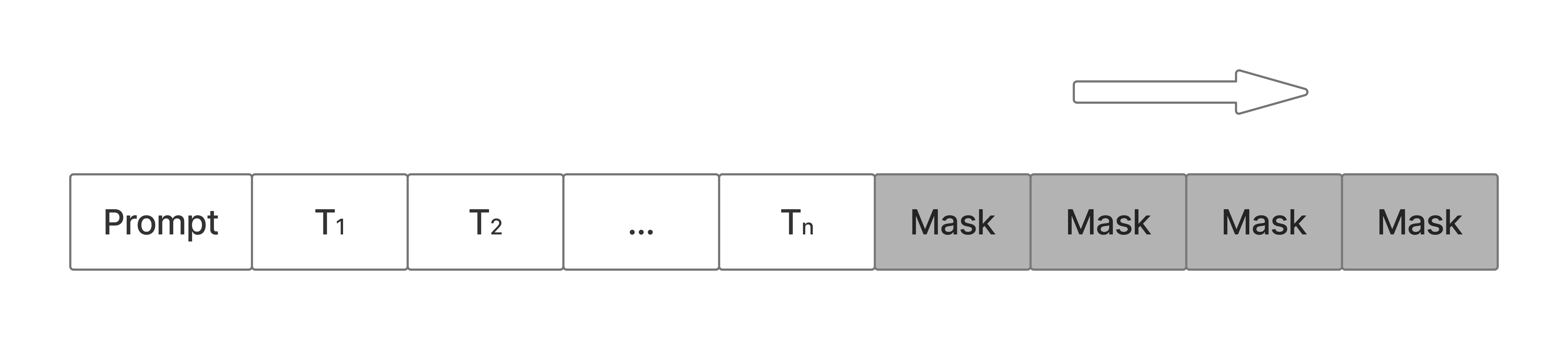}
	\caption{Semi-AR Generation (Block size=4)}
	\label{fig:fig5}
\end{figure}

\paragraph{Blockwise Reverse-Order Training}
During training, we replace a block of tokens with \texttt{[MASK]} tokens in a backward order, starting from the end of the sequence. The prediction of the tokens within this block serves as the supervision signal. Afterward, we remove the content of this block and move one block further toward the beginning, masking the corresponding tokens. This process continues until the remaining sequence is shorter than a block; if there are still residual tokens, we allocate a block from the leftmost side and let the model generate it unconditionally.

This training scheme better aligns with our generation approach and is more efficient, as each block only needs to be trained for $2^4-1$ cases (if block size = 4), which is easier to achieve.

\section{Conclusion}
In this work, we have provided a theoretical and empirical analysis of the limitations of mask diffusion language models: their predictions collapse to marginal distributions, parallel sampling lacks joint coherence, and the generation process is essentially autoregressive, which misaligns with the training procedure.

We also proposed a training and inference framework that better aligns with current mask diffusion language models, although it does not fundamentally enhance their ability for parallel generation or the effective use of bidirectional attention.

Our findings suggest that future work should investigate diffusion approaches that enable genuine parallel generation and effective bidirectional attention, while remaining computationally efficient in training.

For more information, please visit our website:
\begin{center}
  \url{https://whaletech.ai/}
\end{center}

% ===== 在 conclusion 后插入 =====

\clearpage                 % 结论结束后换页

% 若正文本是双栏，想让 Team 页单栏显示，则解开下一行
% \onecolumn

{\Huge\bfseries\color{whaleTeal} WhaleTech.ai Team\par}
\vspace{1em}

\textbf{Core Contributors:}
\begin{plainlist}
	\item Haocheng Sun
	\item Cynthia Xin Wen
	\item Edward Hong Wang
\end{plainlist}

\textbf{Research (alphabetically by first name):}
\begin{plainlist}
	\item B.Y. Wei
	\item J.Y. Shi
	\item T. Zhou
	\item Z.X. Yang
\end{plainlist}

\textbf{Operation (alphabetically by first name):}
\begin{plainlist}
	\item J.F. Liu
	\item X.W. Zhang
\end{plainlist}

\clearpage                 % Team 页结束 → 新开一页放参考文献

% 若之前切到单栏，这里想把参考文献恢复成双栏，则解开下一行
% \twocolumn

% ===== 参考文献 =====
% （bibtex 流程）
\bibliographystyle{unsrt}    % 或者 plainnat 等，与期刊要求一致
\bibliography{references}    % 你的 .bib 文件名

\end{document}